\documentclass[%
 aip,
 amsmath,amssymb,
 reprint,%
floatfix
]{revtex4-1}

\usepackage{graphicx}
\usepackage{dcolumn}
\usepackage{bm}

\usepackage[utf8]{inputenc}
\usepackage[T1]{fontenc}
\usepackage{mathptmx}

\begin{document}

\preprint{AIP/123-QED}

\title[Title]{Cognitive simulation models for inertial confinement fusion: \\ Combining simulation and experimental data}

\author{K. D. Humbird}
 \email{humbird1@llnl.gov}
\affiliation{ 
Lawrence Livermore National Laboratory,
7000 East Ave, Livermore CA, USA, 94550
}

\author{J. L. Peterson}
\affiliation{ 
Lawrence Livermore National Laboratory,
7000 East Ave, Livermore CA, USA, 94550
}

\author{J. Salmonson}
\affiliation{ 
Lawrence Livermore National Laboratory,
7000 East Ave, Livermore CA, USA, 94550
}

\author{B. K. Spears}
\affiliation{ 
Lawrence Livermore National Laboratory,
7000 East Ave, Livermore CA, USA, 94550
}

\date{\today}

\begin{abstract}
The design space for inertial confinement fusion (ICF) experiments is vast and experiments are extremely expensive. Researchers rely heavily on computer simulations to explore the design space in search of high-performing implosions. However, ICF multiphysics codes must make simplifying assumptions, and thus deviate from experimental measurements for complex implosions. For more effective design and investigation, simulations require input from past experimental data to better predict future performance. 

In this work, we describe a cognitive simulation method for combining simulation and experimental data into a common, predictive model. This method leverages a machine learning technique called ``transfer learning'', the process of taking a model trained to solve one task, and partially retraining it on a sparse dataset to solve a different, but related task. In the context of ICF design, neural network models trained on large simulation databases and partially retrained on experimental data, producing models that are far more accurate than simulations alone. 

We demonstrate improved model performance for a range of ICF experiments at the National Ignition Facility, and predict the outcome of recent experiments with less than 10\% error for several key observables. We discuss how the methods might be used to carry out a data-driven experimental campaign to optimize performance, illustrating the key product – models that become increasingly accurate as data is acquired. 
\end{abstract}

\maketitle

\section{\label{sec:intro} Introduction}
Inertial confinement fusion experiments seek to produce high neutron yield by compressing a small spherical capsule filled with deuterium and tritium, creating conditions that are favorable for fusion reactions~\cite{icf,Lindl}. The experiments are complex and expensive; to determine the experimental setup that is appropriate to test a particular hypothesis or achieve a certain goal, researchers rely heavily on computer models of the system. Due to the complex multiphysics nature of ICF, the computer simulations necessarily make many approximations and simplifying assumptions to reduce the computational cost. While highly predictive for many classes of experiment, simulations deviate from measurements for the most challenging implosions. For more effective design and investigation, simulations require input from past experimental data to better predict future performance. 

In this work, we present a data-driven method for calibrating simulation predictions using previous experiments, creating a model that predicts the outcome of new experiments with higher accuracy than simulations alone across a broad range of design space. The model is a deep neural network trained on simulations, then ``transfer learned'' with experimental data. In section ~\ref{sec:nn}, we provide a brief introduction to neural networks and transfer learning. In section ~\ref{sec:tlicf}, we discuss the challenges of applying traditional transfer learning techniques to laser indirect drive ICF experiments and present an alternative approach to transfer learning. In section ~\ref{sec:tlnif}, we apply the proposed transfer learning technique to National Ignition Facility (NIF) data and demonstrate the predictive capability of the transfer learned model for recent NIF experiments.

\section{\label{sec:nn} Neural networks and transfer learning}
Transfer learning is a popular technique in the machine learning community that enables accurate models to be built with limited size datasets. In the following subsections, we give a brief overview of neural networks and transfer learning, but refer the reader to references for in-depth articles on the machine learning techniques~\cite{TL,TL1,TL2,TLimages,TLimages1,TLimages2,generalTL}.

\subsection{Neural networks}
\begin{figure}
\includegraphics[width=0.25\textwidth]{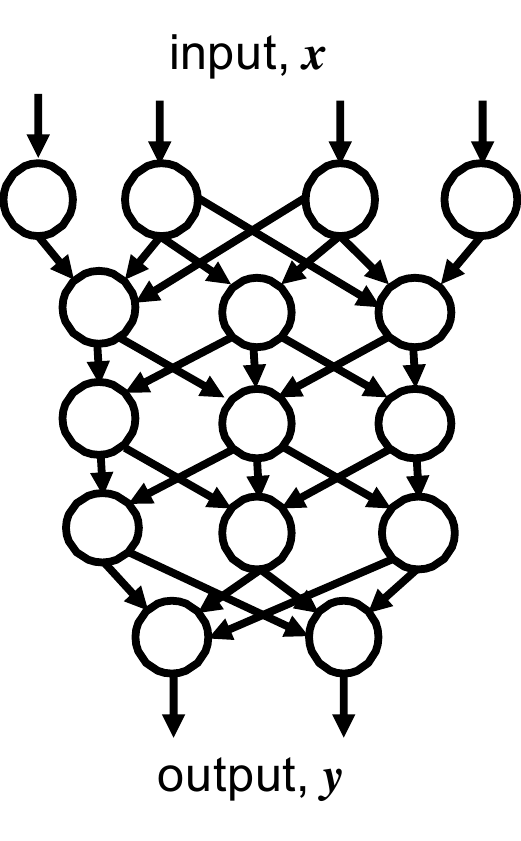} 
\caption{\label{fig:nn} Illustration of a neural network; inputs are mapped to outputs via a series of nonlinear transformations through the layers of the network. The model is trained by learning from a supervised dataset of the inputs and corresponding outputs. }
\end{figure}

Neural networks are popular machine learning models that excel at a variety of tasks~\cite{nnphysics,nnphysics2,rnn,cnn,djinn}. Mathematically, a neural network is quite simple -- it is a series of nested nonlinear functions that are fit to data. An illustration of a simple neural network used to solve a regression task is shown in Figure \ref{fig:nn}. For regression problems, data consists of several samples of independent ``input'' variables, X, and the corresponding dependent ``output'' variables, Y. 

\begin{figure}
\includegraphics[height=0.4\textwidth]{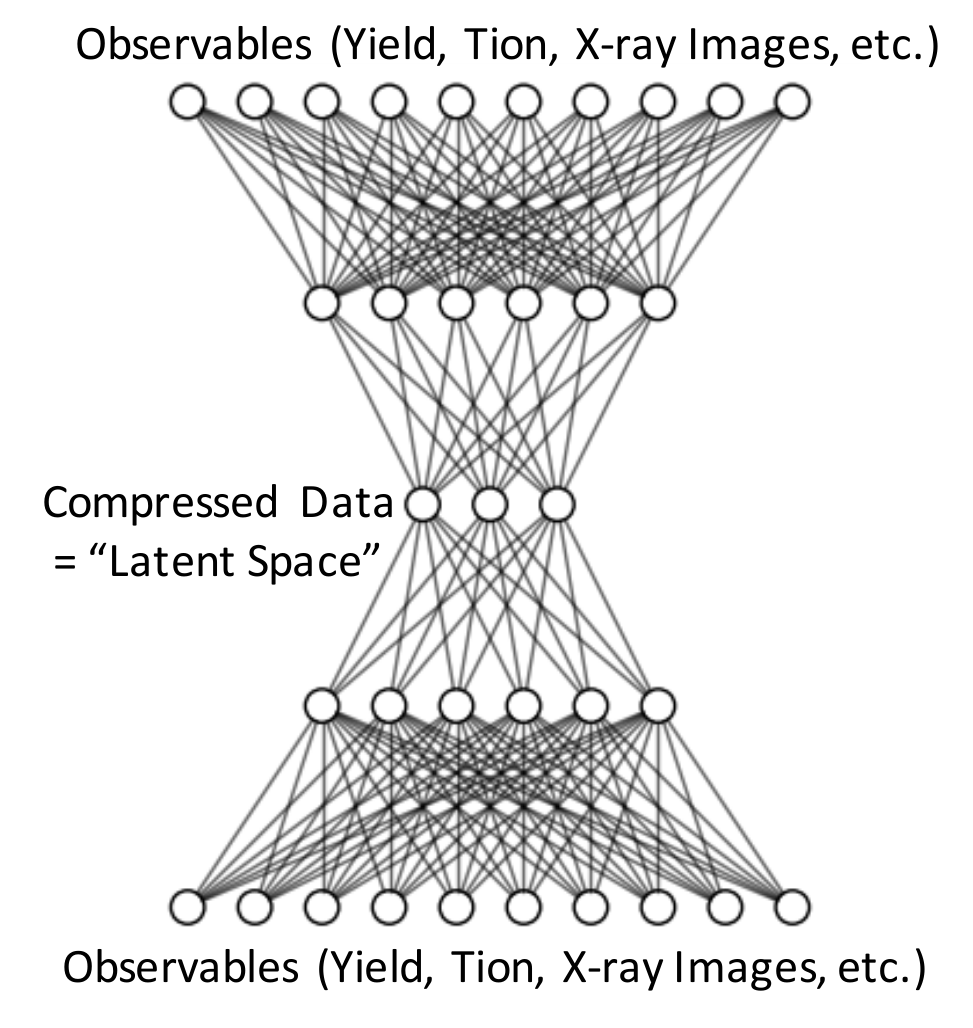} 
\caption{\label{fig:ae} Illustration of an autoencoder neural network. Autoencoders take as input to the model a large set of correlated data, and non-linearly compress this information into a lower dimensional latent space via the encoder portion of the model. The decoder then expands the latent space vector back to the original set of data; the model is trained by minimizing the reconstruction error.}
\end{figure}

The inputs are fed into the first layer of the neural network in the case of a regression problem. The input vector is multiplied by an array of ``weights'' and the product is added to a vector of ``biases''. The resulting vector is transformed by a nonlinear ``activation'' function, the output of which is then used as the input to the next layer. This series of matrix operations and nonlinear transformations continues until the output layer of the network, which aims to produce the corresponding outputs variables for the given inputs. In the context of ICF, the input vector might contain the capsule geometry and laser pulse, while the output vector might include neutron yield, ion temperature, and areal density. The neural network weights and biases are often randomly initialized, and then adjusted to minimize the error between the output layer predictions and the true output. The network learns the mapping between the inputs and outputs by observing a large set of samples with varying inputs and the corresponding outputs and adjusting the weights and biases until the prediction error is minimized. For further details on neural network training, we refer the reader to the references~\cite{deeplearning_hintonlecun}. 
Specialized neural network architectures are used to accomplish a variety of tasks beyond regression. A type of neural network that will be used in this work is an autoencoder~\cite{autoencoder}. Autoencoders are networks often used for data compression; a standard autoencoder architecture is shown in Figure \ref{fig:ae}. 

Autoencoders typically (but do not always) have an hourglass-like shape, starting with a wide input layer then reducing in width each layer until a bottleneck layer is reached; the second half of the model then often mirrors the first half, progressively increasing in width until reaching the output layer, which is identical to the input layer. The autoencoder is trained by minimizing the reconstruction error between the input and output layer, while forcing the data to go through the low dimensional bottleneck layer, which is referred to as the ``latent space''. The latent space is essentially a compressed representation of the data input to the neural network. A common use of autoencoders is to compress image data -- the network learns correlations between pixels and leverages these relationships to develop a latent space that represents the fundamental features of the image. Autoencoders have recently been applied to ICF relevant problems for efficient opacity calculations in radiation hydrodynamics simulations~\cite{gilles}. 

\subsection{Transfer learning}
Transfer learning is a machine learning technique in which a model trained to solve one task is partially retrained to solve a different, but related task, often one for which there is insufficient data to train a model from scratch. A common use of transfer learning is object recognition; large neural networks are trained to recognize a variety of generic objects using databases of millions of images. Transfer learning is used to retrain a small portion of this model -- often the last few layers of the neural network -- to identify very specific types of objects, such as an aircraft type or breed of dog. The idea behind this methodology is that networks are observed to decompose images such that features get more detailed as the depth of the network progresses. Early layers in the model may focus on lines, and basic shapes, while deeper layers learn to recognize details more specific to the object being labeled, such as eyes or coat patterns. By modifying just the last few layers, which focus on details that differentiate one similar image from another, one can take advantage of layers in which the model learned generic image features, without having to provide copious data and training time required for the model to learn such information.

\section{\label{sec:tlicf}Transfer learning for inertial confinement fusion}
In the context of modeling inertial confinement fusion experiments, the simulations provide a map, or response surface, of implosion performance as different design parameters are varied. Simulations are an approximation to the experiments, and thus the map they provide is also an approximation, and might have errors due to unresolved, or inaccurate physics in certain regions of design space. A neural network trained on a large database of ICF simulations provides a continuous map of implosion performance based on the model used to generate the data. There are several techniques that have been explored to make the simulation map a better representation of the true experimental map. Drive multipliers, mix models, and other physics-informed adjustments that are made in order to bring simulation models into consistency with specific experiments is a common method for improving simulation accuracy in the region immediate surrounding the experiment~\cite{Clark:2008fj}. Bayesian calibration~\cite{JimBayesian,coda} is an extension of this technique that provides adjustments to the simulation as a function of the design parameters that applies across a campaign of similar shots. Transfer learning has also been explored for ICF~\cite{humbirdTL,kustowskiTL}, but in all of these examples the models are limited to small design spaces of no more than nine input parameters, making the applicability of the model quite restrictive. Statistical methods for calibrating ICF simulations to experiments have been successfully applied to direct drive experiments~\cite{lle_nature}, though the technique is limited to power-law relationships between simulation outputs and experimental observables. 

In this work, we present a new approach to transfer learning for creating predictive models of ICF experiments that is not limited to compact design spaces. In the following section we summarize the traditional approach to transfer learning that maps from design parameters to experimental observables, and discuss the challenges of applying this technique to indirect drive ICF experiments carried out at the National Ignition Facility. In section ~\ref{subsec:tl1}, we introduce a novel approach to transfer learning for ICF based on autoencoder networks, essentially creating a corrective transformation to simulation predictions that make them more consistent with experimental reality. The resulting model corrects the simulations, which have average prediction errors of 110\%, and produces predictions with an average prediction error of less than 7\%. 

\subsection{\label{subsec:tl1}Transfer learning for indirect drive ICF}
In previous work, transfer learning has been used to create predictive models for ICF experiments for specific experimental campaigns. In those studies, a neural network trained to map from design parameters (experimental inputs) to measurable outputs (experimental observables) are initially trained on large databases of simulations chosen to span the space of the experimental campaign. The small set of experimental data points are then used to retrain a small portion of the simulation-based neural network (typically the last layers of the network) to adjust the model from predicting simulation outputs to predicting experimental outputs. 

An advantage of this model is that optimizing design parameters for experimental performance can be done quickly -- the neural network can be evaluated millions of times in minutes to search the design space spanned by the data for high performing regions of the design space. A disadvantage of this approach is that the design space that can be explored is limited to about ten parameters, otherwise it becomes infeasible to create a database of simulations that adequately fill the space, and the density of experimental data points becomes extremely sparse. Since the models are limited to low-dimensional input or design spaces, the number of experiments that can be included in the model is also limited. ICF experiments have myriad design parameters and existing databases, such as the database of ICF experiments executed at the National Ignition Facility (NIF)~\cite{NIFMiller}, sparsely fill this extremely high dimensional space. It is difficult to find more than a dozen NIF experiments that are confined to a 10 dimensional volume design space. This means many experiments are unused in the prior work on transfer learned models. 

An additional challenge that transfer learning poses for indirect drive ICF is that our highest fidelity simulations only model the capsule. The capsule simulations take a radiation drive produced by an integrated simulation that includes the hohlraum, which is often not predictive of experimental observables, and adjustments are applied to the capsule drive in order to reproduce experimental measurements for several quantities of interest (QOI). In short, the consequence of this technique is the parameters that are input to the experiment are different than the parameters input to the simulation, thus modifying the input to output mapping, as done previously~\cite{humbirdTL}, is not applicable to indirect drive experiments modeled by capsule simulations. Kustowski et al~\cite{kustowskiTL} perform traditional transfer learning for indirect drive by first inferring capsule inputs for each experiment via Bayesian inference; transfer learning enables them to predict observables that could not be matched simultaneously via the initial inference step. 

An alternative to Bayesian inference of capsule inputs prior to traditional transfer learning is to generate ensembles of hohlraum simulations instead of capsule-only simulations. This is computationally intensive, however, and in practice is limited to design spaces of approximately five dimensions in order to train an accurate neural network emulator of the simulations. This makes the challenge of experimental data sparsity more difficult to handle, as there are few NIF experiments confined to such a low dimension parameter space. 

\subsection{Transfer learned auto-encoders for ICF}
Transfer learning using the traditional technique for indirect drive ICF is limited to exploring small design spaces, as discussed in the previous section. Currently, the sampling density of NIF experiments does not enable us to use the technique outlined in the thought experiment of the last section on real experimental data. In order to use transfer learning for the current NIF data, novel transfer learning techniques are required. 

While the sampling density in the input space of NIF experiments is extremely sparse, many of the NIF ignition implosions (those that contain deuterium and tritium, or DT, fuel) aim to achieve high performance, and thus achieve very similar output conditions. So while each campaign might be confined in its own volume of design space, many of these campaigns have comparable performance, and thus span a compact volume of output space. The sampling density of the output space is thus quite high, with several dozen experiments achieving similar neutron yield, ion temperature (T$_{ion}$), down scatter ratio (DSR), etc. Instead of having transfer learned models learn the correct mapping from design parameters to outputs, instead we re-frame the problem to learn a correctional transformation to our simulation outputs, such that we produce predictions that are more consistent with the outputs measured in the experiments. Similar work has been done with power law models that connect simulation outputs to experimental outputs for Omega ICF experiments ~\cite{lle_nature}; we take a neural network-based approach to allow for more flexibility in mapping between simulations and experiments.

\begin{figure}
\includegraphics[width=0.5\textwidth,angle=-90]{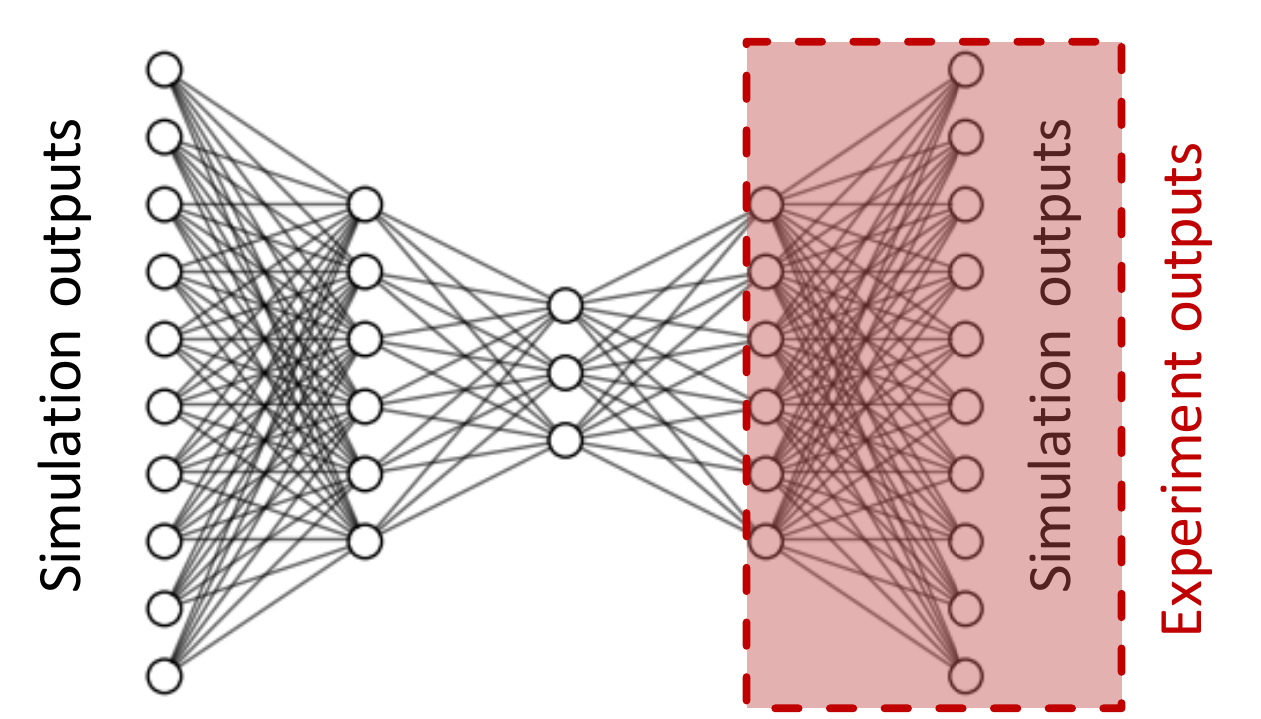} 
\caption{\label{fig:tlaecart} Illustration of how an autoencoder is transfer learned from decoding to experimental outputs rather than simulation outputs. }
\end{figure}

To learn the mapping from simulation outputs to experimental outputs, we transfer learn an autoencoder. The idea is illustrated in Figure~\ref{fig:tlaecart}. First, an autoencoder is trained on simulations to learn how observables are correlated with one another, forming a latent space representation of the simulation outputs. The decoder leverages the correlations to expand the latent space back to the original vector of simulation observables put into the autoencoder. However, we can leverage the idea of transfer learning such that, rather than decoding to the simulation observables, we decode to the experimental observables. The model will adjust the relationships between observables implied by the simulation data to instead be more consistent with what is observed across a variety of experiments. The transfer learned autoencoder essentially provides us a mapping from simulation outputs to experimental outputs. This corrective transformation can be applied to any set of future simulation predictions, giving the researcher a more realistic expectation of how the proposed experiment will perform based on discrepancies between previous experiments and their corresponding simulations.

\begin{figure*}
\includegraphics[width=0.85\textwidth]{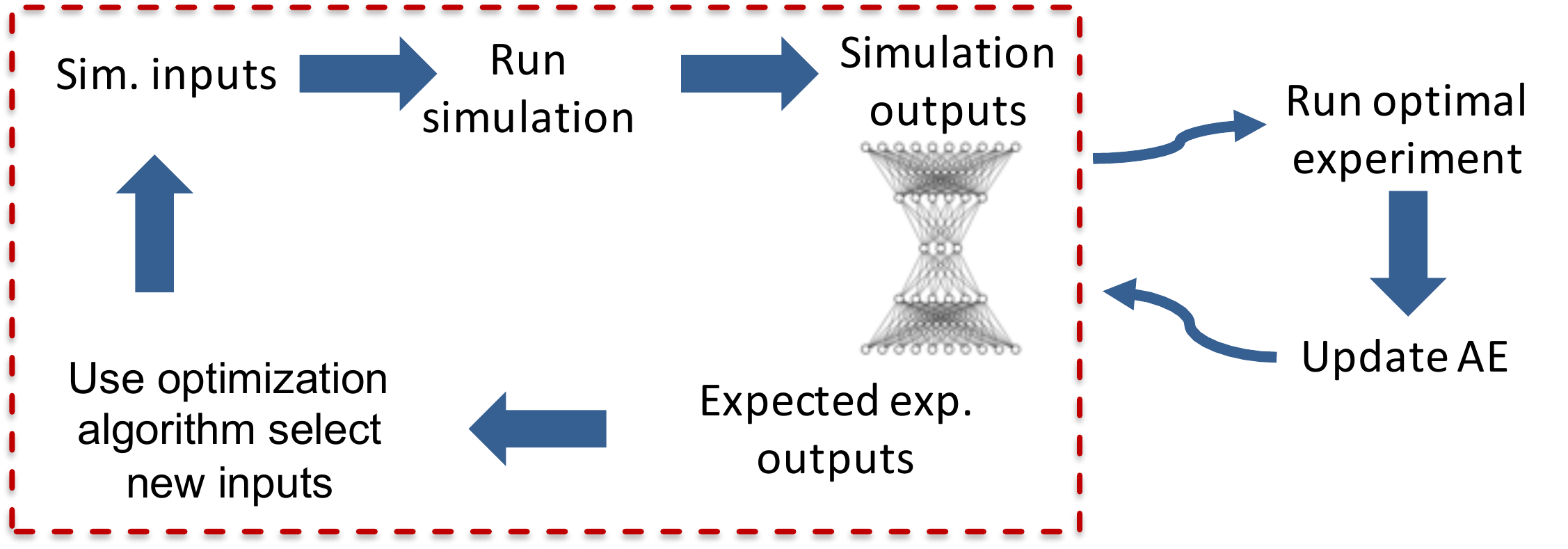} 
\caption{\label{fig:wfae} Proposed workflow for design experiments using a data-informed transformation that is applied to simulation outputs. }
\end{figure*}

A benefit of transfer learning is the model can be updated iteratively as experimental data is acquired, getting more accurate over time and covering a broader range of output space as our experiments diverge in performance from the existing NIF database. This workflow is illustrated in Figure~\ref{fig:wfae}: a researcher comes up with a design and runs a simulation, producing a set of simulation outputs. The outputs are transformed via the transfer learned autoencoder to produce data-informed predictions of how the experiment will perform. The researcher can optimize their design using the corrected predictions, then execute the optimal implosion. The data collected in this experiment is then fed back into the transfer learned autoencoder, increasing the model's predictive capability as more experimental data is acquired.  

In the next section, we demonstrate the first few steps of this proposed approach to creating predictive models of NIF experiments leveraging existing NIF data. We use a simulation database and data from previous NIF experiments to create a transfer learned autoencoder that accurately maps between simulation outputs and experimental measurements for a wide variety of NIF DT experiments.

\section{\label{sec:tlnif}Transfer learned autoencoders for NIF experiments}
To train an autoencoder to map from simulation outputs to experimental outputs, two datasets are needed -- a simulation database on which the autoencoder is initially trained in order to learn how observables are correlated with one another -- and a database of experiments and their corresponding simulation predictions for transfer learning. For the underlying simulation database on which the autoencoder is first trained, we use a set of 160,000 2D Hydra capsule simulations that have been generated over years of research~\cite{bigfoot,nora,JimBayesian}. These simulations span a broad output space that encompasses a large number of NIF experiments. 

The supervised set of simulation predictions and experimental observations for a given design are generated using the HyPyD~\cite{hypyd} framework for Hydra~\cite{hydra}, enabling us to create a set of database of predictions using a common, unaltered model (no drive multipliers, mix models, etc.). To keep the computational cost low for this proof of principle, we use 1D hohlraum simulations to create predictions for several observables that are available from the experiments. Future work will focus on improving the simulation predictions using 2D hohlraums; since the discrepancy between 2D hohlraums and reality is less than that between 1D and reality, we expect the results to improve with better modeling fidelity. 

The experiment database is a collection of NIF DT shots executed in the last three years, for which the relevant observables are readily available. This totals 47 experiments that span a variety of campaigns, including Bigfoot~\cite{bigfoot}, HDC~\cite{hdc,hdc2}, 2-shock~\cite{2shock}, HighFoot~\cite{highfoot,bf2}, Hybrid B~\cite{hybridb1,hybridb2,hybridb}, Hybrid E~\cite{hybride,hybride2}, and the I-raum~\cite{iraum}. The observables that the model predicts include the gamma bang time, gamma burnwidth, the DT neutron yield and ion temperature, the DD yield and ion temperature, and the down scatter ratio. 

The autoencoder is a fully connected neural network trained with the open source software TensorFlow~\cite{tensorflow2015}. The architecture starts with seven inputs, expands to two layers with ten neurons each, then maps into a latent space of five neurons, with rectified linear unit activation functions~\cite{relu} applied to each layer. The decoder mirrors the encoder network. The model is trained using the Adam optimizer~\cite{adam} with a mean squared error loss function and a learning rate of 0.01 for 800 epochs with a batch size of 300 simulations. The autoencoder can reconstruct the simulation observables with explained variance scores of over 0.9 for all observables. 

Transfer learning the autoencoder is performed iteratively, adding experiments one at a time in chronological order. With each additional experiment, we update the model by retraining the final two layers of the decoder with a learning rate of 0.001 for 300 epochs with a batch size of one. Each time we add an experiment, we start with the same initial simulation-based autoencoder before transfer learning on the experimental data. The hyperparameters are chosen to ensure convergence of the cost, with a batch size of one to accommodate adding experiments one at a time. Predictive performance is evaluated with each added experiment by computing prediction error on a holdout test dataset. The test dataset includes the seven most recent experiments in the database (including Hybrid E, 2shock, and I-raum experiments); these experiments are never used in the training process, they are strictly for validation of the model. In summary, each time we add a new experiment we transfer learn the initial simulation-based autoencoder using the experimental data, and validate the model's predictive performance by predicting the outcome of the seven most recent experiments in the database.

\begin{figure}
\includegraphics[width=0.45\textwidth]{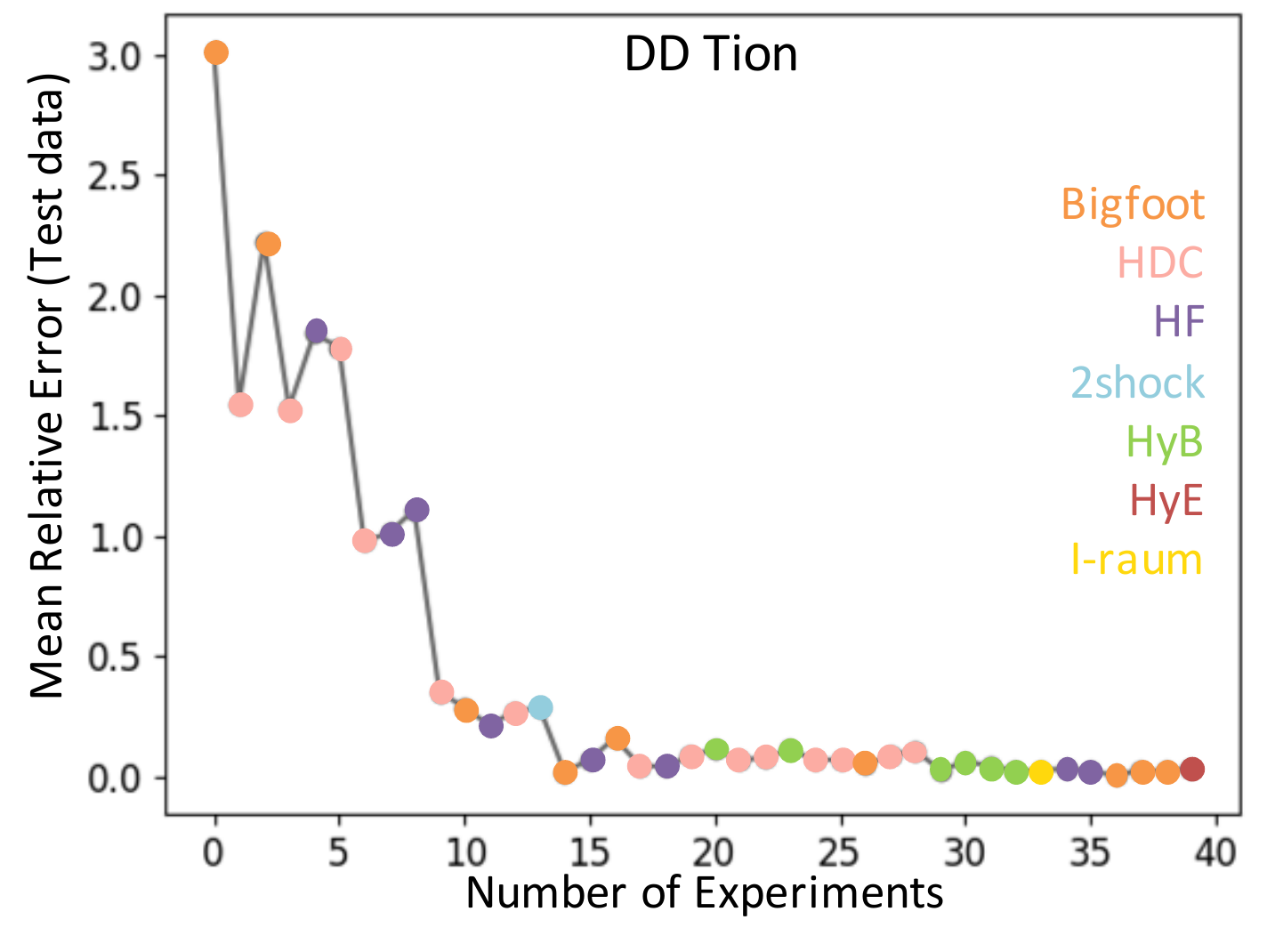} 
\caption{\label{fig:dd} Prediction error on the validation set for the DD ion temperature decreases as experimental data is acquired. The color of the point indicates the experimental campaign for that data point, illustrating how the error changes as the model sees a wider variety of design space.}
\end{figure}

\begin{figure}
\includegraphics[width=0.45\textwidth]{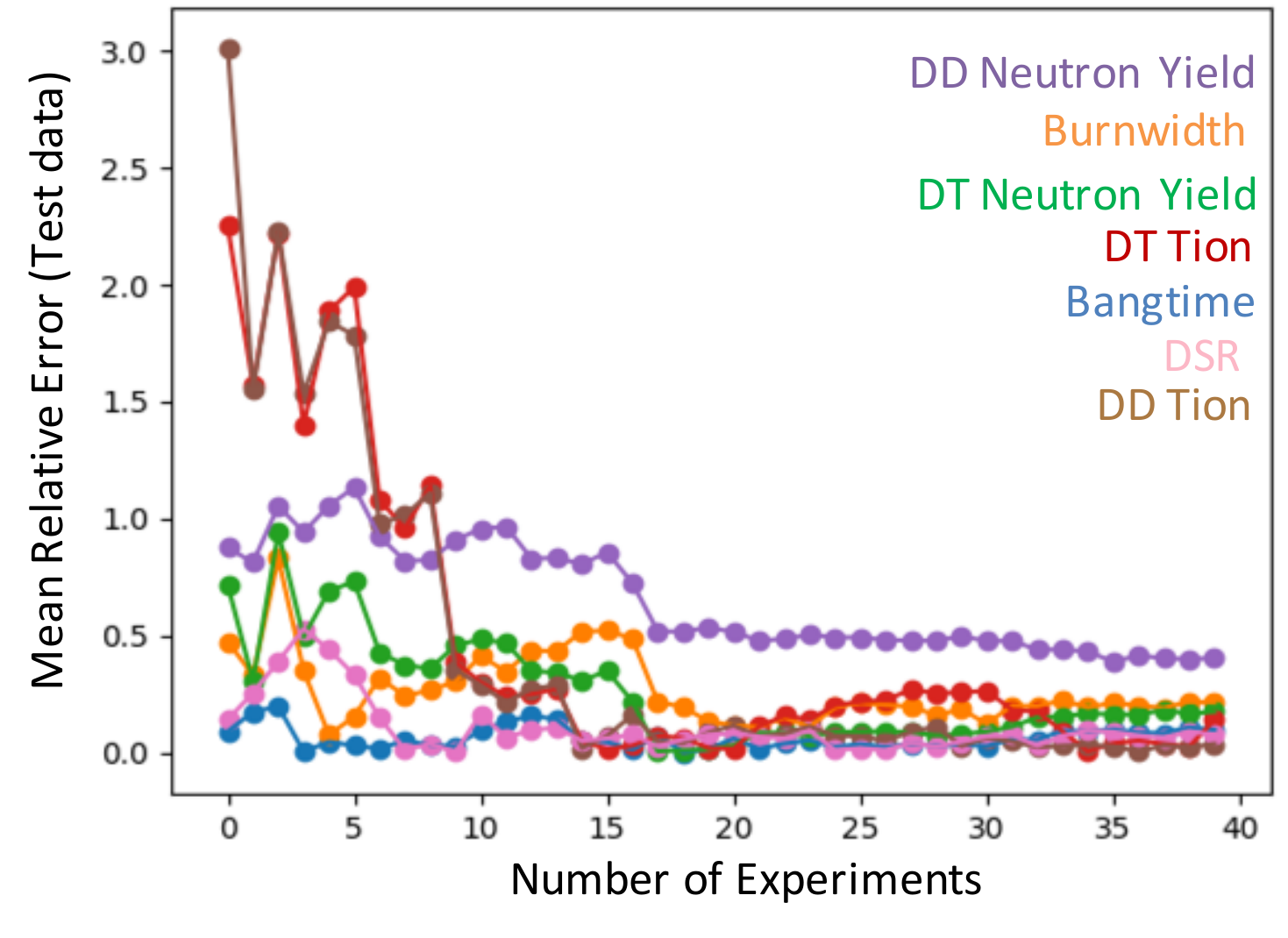} 
\caption{\label{fig:error} The prediction error on the validation set decreases for all quantities of interest as experimental data is acquired. The model error converges around 15-20 experiments. }
\end{figure}

Figure~\ref{fig:dd} shows how the prediction error changes as experiments are added to the transfer learning process for one observable measured in the DT experiments (the deuterium-deuterium, or DD, ion temperature). The points are colored by experimental campaign, illustrating the trajectory of the model's exploration of design space. In Figure~\ref{fig:error}, we are plotting all seven observables prediction error as a function of the number of experiments included in transfer learning, this time colored by observable. As expected, the error starts high for most observables and gradually decreases, noisily at first, to a minimum prediction error after about 15-20 experiments. This model has relatively low capacity due to the small architecture of the neural network, thus we expect the convergence in prediction error to be a result of limited model capacity in addition to commonality in the experiments performance for these seven observables. We expect if a large array of observables are included in this model (such as X-ray or neutron images) the model would continue to learn as it acquires experimental data, both because the data is richer and the autoencoder architecture would naturally expand. 

\begin{figure*}
\includegraphics[width=0.95\textwidth]{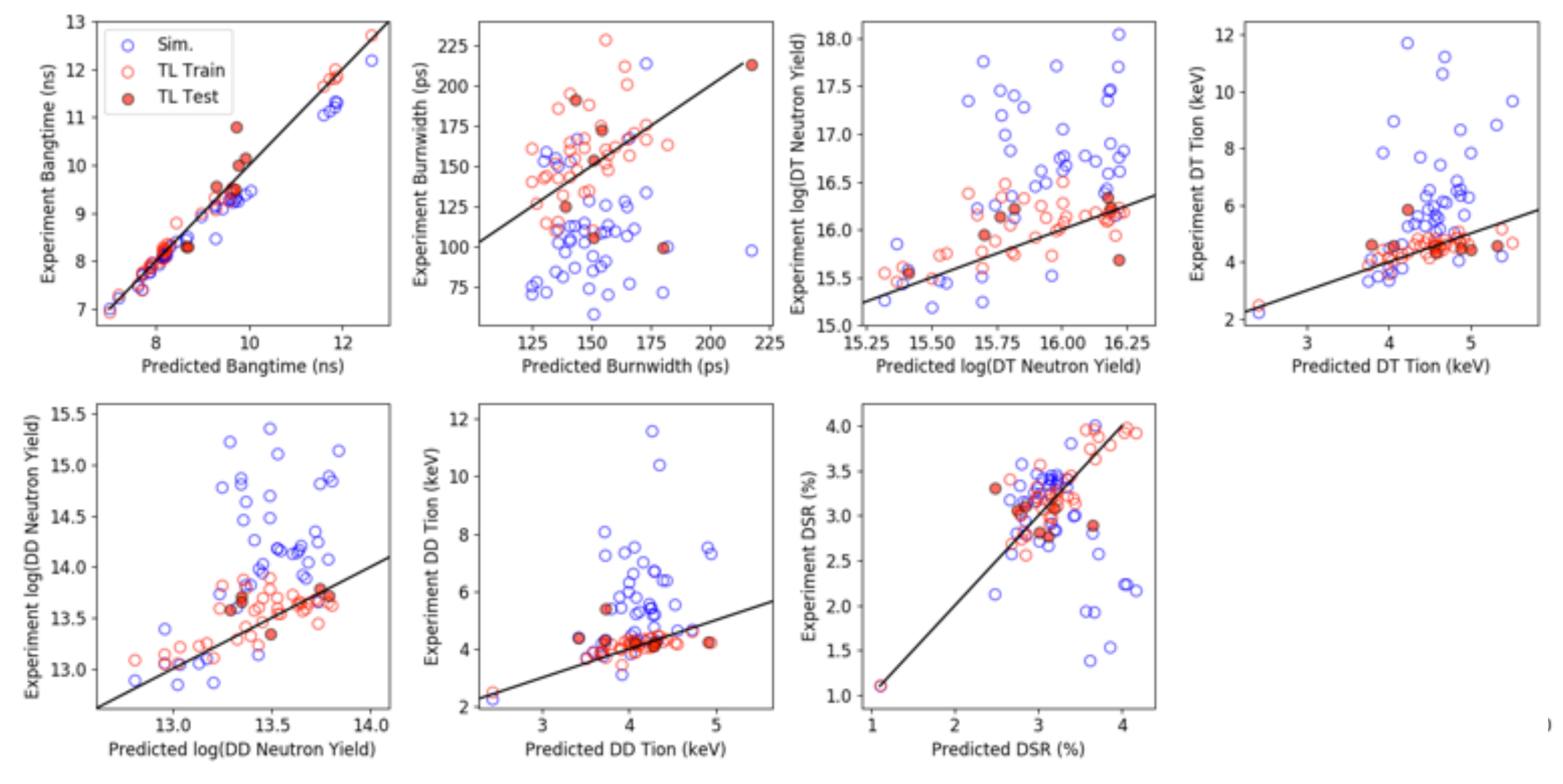} 
\caption{\label{fig:avp} Experimental measurements for seven quantities of interest plotted against the model predictions. The simulation-based predictions are shown in blue, the transfer learned predictions in red, with the seven-shot validation set distinguished by the black outline. The black diagonal line is the y=x line; points falling along this line are perfectly predicted. The transfer learned model is significantly more accurate than the simulation at predicting all quantities of interest. }
\end{figure*}

In Figure~\ref{fig:avp}, the actual values for each quantity of interest are plotted against the predictions after transfer learning on all 40 data points. The blue points are predictions from the simulation alone, red is the training set for transfer learning, and red with black outlines are the seven validation shots predicted by the transfer learned model. The transfer learned model predicts the seven quantities of interest with much higher accuracy than the simulation for the training and validation data. Transfer learning reduces mean relative prediction error on the test data from 9\% to 4\% for bangtime, 50\% to 14\% for burnwidth, 73\% to 3\% for log DT neutron yield, 200\% to 7\% for DT ion temperature, 87\% to 7\% for log DD neutron yield, 300\% to 7\% for DD ion temperature, and  14\% to 8\% for the downscatter ratio, after retraining on 20 experiments.

\subsection{Future improvements}
The previous section presents a proof of concept for using transfer learned autoencoders to map from simulation outputs to experiment observables for NIF indirect drive DT experiments. The results are promising, producing models that predict validation experiments with less than 10\% error for most quantities of interest. Fortunately, there are still many steps that can be taken to improve the accuracy of the model further. First, the underlying simulation database on which the model is trained can be expanded to include hohlraum simulations; the relationships between observables according to 2D capsule simulations are different than the relationships predicted by hohlraums due to limitations in the drive asymmetries explored in the capsule simulations. These relationships are also different than what would be observed in 3D simulations, however 3D simulations, in capsule or hohlraum form, are prohibitively expensive. Hierarchical transfer learning~\cite{humbirdTL} could be explored for taking a 2D simulation-based autoencoder, and elevating it by transfer learning the model with 3D simulations, and then subsequently transfer learning on experimental data. 

The 1D hohlraum model used to generate the predictions for each experiment can be readily improved by instead running 2D hohlraum simulations. The 2D hohlraum simulations are significantly more computationally expensive than the 1D model, but it is not infeasible to run a 2D hohlraum simulation for each NIF DT experiment. 

Finally, more NIF experiments can readily be included in the model. A subset of available experiments, consisting of 47 shots is used in this study; another 50-60 shots can be added in future iterations of this work. Additionally, more observables can be included, such as image information from the X-ray and neutron images collected for most experiments; recent work is exploring transfer learning for NIF image data~\cite{kustowskiAPStl} and we expect these efforts can be combined to create autoencoders that accurately predict a wide array of observables for NIF experiments.

\section{\label{sec:concl}Conclusions}
Predictive modeling for indirect drive inertial confinement fusion experiments is challenging; models are often only accurate across a small volume of design space and require tuning due to a number of necessary approximations. In this work, we present a data-driven approach to learning the corrective transformation that must be applied to simulations of NIF experiments in order to make their predictions consistent with observations. This method uses a neural network technique called transfer learning to take simulation based autoencoder which, rather than decoding from a latent space to simulation outputs, decodes to experimental outputs. The result is a nonlinear transformation that is applied to simulation outputs to generate data-informed expected experimental observations. Using inexpensive 1D hohlraum simulations, we are able to create a model that is consistently predictive of NIF experiments across a broad range of campaigns, with less than 10\% prediction error for most quantities of interest. This model is useful for design, enabling researchers to optimize implosion performance based on experimental expectations, rather than simulation predictions alone.

\begin{acknowledgments}
The authors would like to thank Bogdan Kustowski for conversations on transfer learning, and Omar Hurricane and Debra Callahan for their feedback on this work. 

This work was performed under the auspices of the U.S. Department of Energy by Lawrence Livermore National Laboratory under Contract DE-AC52-07NA27344. Released as LLNL-JRNL-817715.
This document was prepared as an account of work sponsored by an agency of the United States government. Neither the United States government nor Lawrence Livermore National Security, LLC, nor any of their employees makes any warranty, expressed or implied, or assumes any legal liability or responsibility for the accuracy, completeness, or usefulness of any information, apparatus, product, or process disclosed, or represents that its use would not infringe privately owned rights. Reference herein to any specific commercial product, process, or service by trade name, trademark, manufacturer, or otherwise does not necessarily constitute or imply its endorsement, recommendation, or favoring by the United States government or Lawrence Livermore National Security, LLC. The views and opinions of authors expressed herein do not necessarily state or reflect those of the United States government or Lawrence Livermore National Security, LLC, and shall not be used for advertising or product endorsement purposes. 
\end{acknowledgments}

\section*{Data Availability Statement}
Data subject to third party restrictions.

\bibliography{aipsamp}

\end{document}